\documentclass[conference]{IEEEtran}
\IEEEoverridecommandlockouts
\usepackage{cite}
\usepackage{amsmath,amssymb,amsfonts,bm}
\usepackage{algorithmic}
\usepackage{graphicx}
\usepackage{textcomp}
\usepackage{xcolor}
\usepackage{url}

\def\BibTeX{{\mathbb{R}m B\kern-.05em{\sc i\kern-.025em b}\kern-.08em
    T\kern-.1667em\lower.7ex\hbox{E}\kern-.125emX}}
\begin{document}

\title{Self-attention as an attractor network: transient memories without backpropagation
\thanks{M. N. acknowledges the support of PNRR MUR Project No. PE0000013-FAIR.}
}


\author{\IEEEauthorblockN{1\textsuperscript{st} Francesco D'Amico}
\IEEEauthorblockA{\textit{Physics department} \\
\textit{University of Rome Sapienza}\\
Rome, Italy \\
francesco.damico@uniroma1.it}
\and
\IEEEauthorblockN{2\textsuperscript{nd} Matteo Negri}
\IEEEauthorblockA{\textit{Physics department} \\
\textit{University of Rome Sapienza}\\
Rome, Italy \\
matteo.negri@uniroma1.it}
}

\maketitle

\begin{abstract}
Transformers are one of the most successful architectures of modern neural networks. At their core there is the so-called attention mechanism, which recently interested the physics community as it can be written as the derivative of an energy function in certain cases: while it is possible to write the cross-attention layer as a modern Hopfield network, the same is not possible for the self-attention, which is used in the GPT architectures and other autoregressive models. In this work we show that it is possible to obtain the self-attention layer as the derivative of local energy terms, which resemble a pseudo-likelihood.
We leverage the analogy with pseudo-likelihood to design a recurrent model that can be trained without backpropagation: the dynamics shows transient states that are strongly correlated with both train and test examples. Overall we present a novel framework to interpret self-attention as an attractor network, potentially paving the way for new theoretical approaches inspired from physics to understand transformers.
\end{abstract}

\section{Introduction} 

Transformer neural networks \cite{vaswani_attention_2017} have caught the interest of the physics community because of their connections with simpler but well-understood models, namely Hopfield networks \cite{ramsauer_hopfield_2021, hoover_energy_2023} and pseudo-likelihood methods \cite{rende_mapping_2024}.

The core idea of this connection is to interpret each layer of a deep feed-forward network as the discrete-time dynamics of some energy system, by writing the output of such layer as the derivative of an energy function. This scheme is not immediately useful when the layers are considered to be different, but when we interpret a deep network as a recurrent network that repeats the same layer (or sets of layers), the question whether the dynamics has attractors and fixed points naturally arises. 

The idea that deep networks would posses attractors has been explored in \cite{bai_deep_2019, radhakrishnan_overparameterized_2020}, showing promising numerical result.
Another evidence of the feasibility of this approach comes from the architecture of Alphafold \cite{jumper_highly_2021}. A part of their pipeline consists in repeating a specific block of layers a random number of times during training, arguing that such procedure improves the accuracy (they call \emph{recycling} such procedure).

An important progress for this framework was made in \cite{ramsauer_hopfield_2021}, where the authors recognized that the cross-attention layer can be written as the discrete-time dynamics of an Exponential Hopfield Model \cite{demircigil_model_2017,lucibello_exponential_2024}. 
The connection with Hopfield networks was strengthened further in \cite{hoover_energy_2023}, where the authors train a repeated block of layers parametrized by Dense Hopfield networks. 
At the same time, Hopfield networks have been shown to be able to generalize the concept of memory to test data \cite{kalaj_random_2024}  and even become generative in the context of diffusion models \cite{raya_spontaneous_2023,ambrogioni_search_2024}, suggesting that this framework could be useful also for theoretical analysis.

The general interpretation of feed-forward layer as derivative of some energy function was discussed in \cite{yang_transformers_2023} with the name of \emph{unfolding}, where the authors describe the challenges of identifying an energy function for the self-attention layer (rather than the cross-attention), which requires a context-dependent calculation of the attention scores.
Furthermore, interpreting self-attention as recurrent updated is much more natural than cross-attention, since the main task of predicting a token given the other tokens in the sequence is by design an auto-associative task.

In this work we connect all these ideas providing a framework to train a self-attention layer as an attractor network.
To do this, we propose a change of perspective from \cite{ramsauer_hopfield_2021}, where the spins have been viewed as the internal degrees of freedom of each token. 
Instead, we consider each token as a vector spin in dimension $d$, as it was suggested in \cite{bal2021deepimplicitattention,bal2021isingisallyouneed,bal2023spinmodeltransformers}. A similar scheme was suggested in \cite{rende_mapping_2024}, where they used Potts variables (that can be seen as a discrete version of vector spins).

In this framework, we can describe the task of predicting a token of a sequence in terms of physics systems: we update the orientation of a vector spin given the local field produced by the others. We will see that the self-attention mechanism modifies significantly the computation of the local field from a fully connected scheme.
This will allow us to write the update of each token as the derivative of a \textit{local} energy function (local in the sense that each spin optimizes its own energy, see section \ref{sec:model_definition}). Then, we exploit an analogy with pseudo-likelihood noted in \cite{rende_mapping_2024} designing a training scheme that avoid backpropagation by minimizing a negative pseudo-likelihood loss on training data (see section \ref{sec:model_training}).

We test the model on two tasks: masked token prediction and denoising. We compare it to more advanced versions of itself, namely a whole transformer block used as recurrent network, and a standard vision transformer (see section \ref{sec:results}).


\section{Self-attention as an attractor network}
\subsection{Model definition}
\label{sec:model_definition}

We consider a sequence of tokens of fixed length $N$. Each token is embedded in a $d$-dimensional sphere $\mathcal{S}^d$ of unitary radius, and we refer to these embedded tokens as vector spins $\mathbf{x}_i\in\mathcal{S}^d$, $i=1,...,N$. 

Each spin evolves with the following dynamics:
\begin{equation}
    \mathbf{x}_i^{(t+1)} = \sum_{j(\neq i)} \alpha_{i\gets j} \mathbb{J}_{ij} \, \mathbf{x}_j^{(t)} + \gamma \mathbf{x}_i^{(t)}
    \label{eq:bareSA_dynamics}
\end{equation}
where the coupling $\mathbb{J}_{ij}$ between tokens $i$ and $j$ is a $d\times d$ matrix that couples the different components of the spins. 
The attention mask $\alpha_{i\gets j}$ is defined as
\begin{equation}
    \alpha_{i\gets j} = \mathrm{Softmax}_{[j]}( \mathbf{x}_i^{(t)} \mathbb{J}_{ij} \mathbf{x}_j^{(t)}),
\end{equation}
where $\mathrm{Softmax}_{[j]}(z_j)=\exp(z_j)/\sum_{j'}\exp(z_{j'})$.
The parameter $\gamma$ can be seen as the relative weight of the signal from other spins with respect to the value of spin $i$ itself. We set $\gamma=1$ troughout this work. Inbetween each iteration, we apply a layer normalization that enforces the spherical normalization of the spins.
In order to compute the error or visualize the predicted image, after the last iteration of the dynamics we need to apply the inverse embedding transformation. In this work we use (untrained) linear embeddings, therefore such inversion is possible.
For a detailed explanation of normalization and embeddings see sec.~\ref{sec:SA_architecture}.

This dynamics of each spin can be written as the derivative of an energy function specific to that spin:
\begin{equation}
    \mathbf{x}_i^{(t+1)}=- \left. \frac{\partial}{\partial{x_{i}}} e_i(\mathbf{x}_i,\{\mathbf{x}_j\}) \right|_{\mathbf{x}_i^{(t)},\{\mathbf{x}_j^{(t)}\}} + \gamma \mathbf{x}_i^{(t)}
\end{equation}
where
\begin{equation}
 e_i(\mathbf{x}_i,\{\mathbf{x}_j\})=-\log \sum_{j(\neq i)} \exp{(\mathbf{x}_i \mathbb{J}_{ij} \mathbf{x}_j)}
 \label{eq:e_i}
\end{equation}
is the energy of token $i$. 
As discussed in \cite{yang_transformers_2023}, the interaction between different tokens is essential to transformers and was not captured by the Hopfield network described in \cite{ramsauer_hopfield_2021}. 

Note that, by defining token-wise energies and deriving those only w.r.t. the corresponding spin, we avoided the additional terms that they find in \cite{hoover_energy_2023}, which make the layer that they consider different from a self-attention layer.

\subsection{Connection to self-attention}
There are two points that need to be considered to interpret this model as a self-attention. Let us write the self-attention update in the notation introduced above:
\begin{equation}
    \mathbf{x}_i^\mathrm{OUT} = \sum_{j(\neq i)}\mathrm{Softmax}_{[j]}(\sum_{k} (\mathbf{x}_i \cdot \mathbf{W}_{k}^Q) (\mathbf{W}_{k}^K \cdot \mathbf{x}_j)) \mathbb{W}^V \mathbf{x}_j 
    \label{eq:actual_SA}
\end{equation}
where the queries and the keys are respectively 
$\mathbf{x}_i \cdot \mathbf{W}_{k}^Q$ and
$\mathbf{W}_{k}^K \cdot \mathbf{x}_j$, the scalar product is intended in the embedding dimension and the index $k$ runs on the so-called internal dimension of the attention.

The first step to write eq.~\ref{eq:actual_SA} as a derivative is to 
define a $d\times d$ coupling matrix as
$
    \mathbb{J}=\sum_{k} \mathbf{W}_{k}^Q \otimes \mathbf{W}_{k}^K.
$
Then, if we force the attention couplings $\mathbb{J}$ to be equal to the values coefficients $\mathbb{W}^V$, we can indeed write $\mathbf{x}_i^\mathrm{OUT}=-\frac{\partial}{\partial{x_{i}}} e_i(\mathbf{x}_i,\{\mathbf{x}_j\})$.

The second step is noticing that in the previous section we used couplings $\mathbb{J}$ that also depend on the spin indices $i,j$. Real transformers would use token embedding that also encode information of the position of the token within the sequence (see \cite{vaswani_attention_2017}). In this work we consider random embeddings (we do not optimize them during training) that do not carry any information on the position of the tokens, and if we did not consider position-dependent couplings $\mathbb{J}$ we would lose fundamental information. The resulting coupling is a $N\times N \times d \times d$ tensor, which makes it very large to store. Nevertheless, given that the model is shallow, we found that the training procedure described in the next section is surprisingly fast.

Note that this setting is restricted to sequences of fixed length, hence the focus on vision transformers. Switching to sequences of variable length would require a proper treatment for the positional encoding.

\subsection{Training procedure}
\label{sec:model_training}

The structure of the token-wise energy contributions in this model resembles the structure of the pseudo-likelihood method used \cite{besag_statistical_1975} to infer the couplings of energy-based models (for a review, see \cite{nguyen_inverse_2017}): instead of maximizing the likelihood of the data, the method proposes to maximize the (log) product of conditional probabilities of each spin given the others. 
Notice that this methods deals explicitly with the challenge of dealing with asymmetric couplings described in \cite{yang_transformers_2023}, since the coupling obtained from the maximization are in general non symmetric.

To apply the same framework to the self attention, let us define the total energy as
\begin{equation}
E({\bf x})=\sum_i e_i(\mathbf{x}_i,\{\mathbf{x}_j\}),
\end{equation}
which allows us to define the loss function as the sum of the energy $E({\bf x})$ evaluated in each data point $\bm{\xi}^\mu$ (embedded in $[\mathcal{S}^d]^N$) of a given training set:
\begin{equation}
    \mathcal{L}(\{\mathbb{J}_{ij}\})=\sum_{\mu=1}^P E(\bm{\xi}^\mu)
    \label{eq:loss}
\end{equation}
Our training procedure consist in finding the value of couplings $\mathbb{J}_{ij}$ that minimize eq.~\ref{eq:loss}. We do it using stochastic gradient descent with a number of regularization precautions, the most important being fixing the norm of $\mathbb{J}_{ij}$ and clipping the gradients. The details of this training are reported in sec.~\ref{sec:SA_training}.

We stress again that our training does not require backpropagation since the model is shallow, nor we impose any symmetry on $\mathbb{J}_{ij}$.

\begin{figure*}[t]
\centering
\includegraphics[width=2\columnwidth]{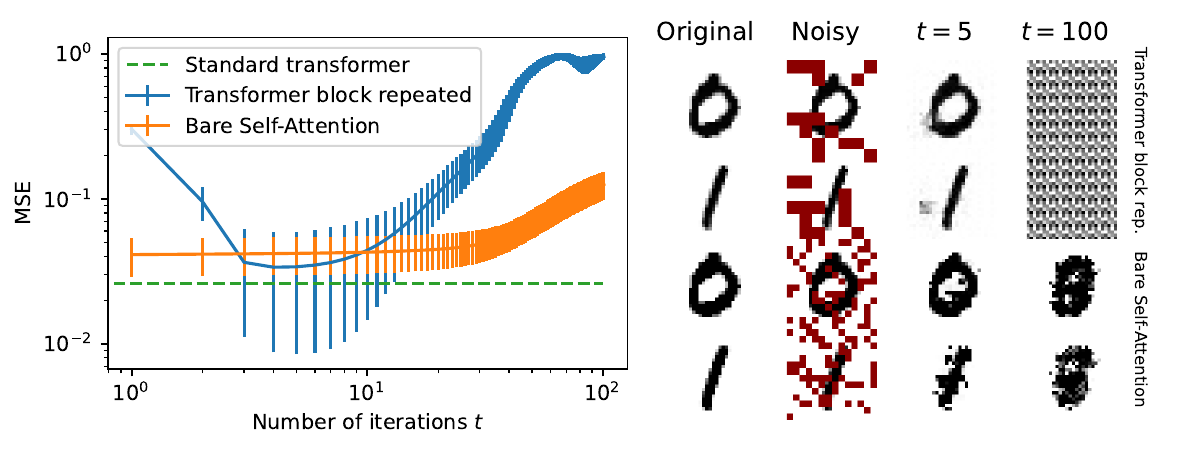}\\
\includegraphics[width=2\columnwidth]{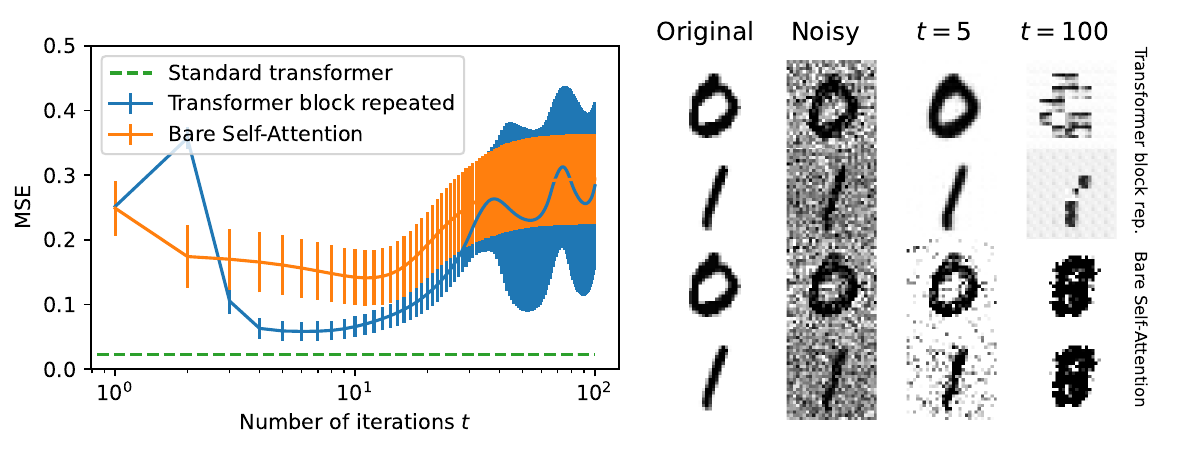}
\caption{
\textbf{Test examples appear as transient states of the dynamics.}
We compare the performance on the test set of various models on two different tasks: in the upper panels we show a masked prediction task, in the lower panels we show a denoising task. Note that the transformers are trained separately for the two tasks, while the self attention is trained in a task-agnostic way.
The plots in the left panels show the mean square error of a prediction after $t$ iterations of the model: the blue lines correspond to a transformer block, the orange to the bare self attention layer and the green dashed to a standard 5-layer transformer (the line is horizontal because we do not repeat this model).
The columns in the right panels show examples of performances of transformer blocks and bare self attention. 
In the masked prediction task we masked $30\%$ of the pixels. In the denoising task we added a Gaussian noise with zero mean and $0.7$ variance on each pixel.
Note that the $t=1$ the denoising task corresponds to the MSE of the corrupted images and $t=2$ is the error after the first iteration. For the masked perdition, the error of the corrupted images is not well defined; therefore, $t=1$ is the error after the first iteration. 
}
\label{fig:results}
\end{figure*}

\section{Results}
\label{sec:results}


In order to test the performance of the Bare Self-Attention recurrent network defined in eq.~\ref{eq:bareSA_dynamics}, we measure its capability of predicting tokens in two settings: the masked prediction task and the denoising task. 

The masked prediction task consist in setting to zero a fraction of tokens of an image from the test set reconstructing of the original image. Results for this task are shown in the top panels of fig.~\ref{fig:results}.

The denoising task is similar to the masked prediction, but we add a fixed level of noise all pixels instead of masking random tokens. Then we rescale the pixel intensities so that the total variance of the noisy image is equal to the variance of the original one. Results for this task are shown in the bottom panels of fig.~\ref{fig:results}.

Since the scope of this work is limited to a proof of concept of a new framework, we keep the benchmarks as simple as possible. In fact, the model does not include trainable embeddings nor the multilayer perceptrons (MLP) after the output of the self-attention layer \cite{vaswani_attention_2017}, which have been found hard to eliminate from transformers blocks without a significant decrease of the performances \cite{he_simplifying_2024}. 

We train the models on 50k images from MNIST, which is a sufficiently simple dataset to allow the strongly model in study to still produce meaningful predictions. Note that the training scheme described in section \ref{sec:model_training} does not depend on the desired task, and therefore we train it once and we test it on the two tasks. These results are shown as orange curves in fig.~\ref{fig:results}.

We benchmark the Bare Self-Attention recurrent network against models that gradually reintroduce the simplified features. 
The first benchmark model is a transformer block that includes the MLP after the Bare Self-Attention layer. During training, the model takes as input the corrupted image and iterates the same transformer block a number of times uniformly distributed between 3 and 7, as it was done in the recycling part of Alphafold \cite{jumper_highly_2021}. We train the model with backpropagation (see sec.~\ref{sec:block_training} for more details). Note that the model need to be trained separately for the masked prediction and the denoising tasks. After training, we measure the reconstruction error as a function of the number of iterations on data from the test set (see blue curves in fig.~\ref{fig:results}). 

The second benchmark model is a 5-layer vision transformer \cite{dosovitskiy_image_2021}
(see green dashed lines in fig.~\ref{fig:results}). 

Both the Bare Self-Attention and the Transformer block show that they "store" test examples as transient states of the dynamics: the behaviour of the Mean Square Error (MSE) is non-monotonic  (orange and blue curves in fig-~\ref{fig:results}), meaning that there is an optimal number of iteration for which the configurations of the spins is maximally correlated with the image from the test set; after that point, the error grows and the configuration goes far away from the initial configuration. 

Note that in the masked prediction task, the Bare Self-Attention reaches an optimal prediction after one iteration, while it need around ten for the denoising case. Also note that the transformer block has a good prediction for 3 to 7 iterations, which are the number that we used during its training.  


The standard vision transformer has the best prediction, and the Trasformer block beats the Bare Self-Attention if given enough iterations. 
One interesting characteristic of the Bare Self-Attention is that it appears to have a global attractor: no matter the task or the specific initialization, the model seems to converge to an image that resembles the average of all MNIST digits. 

Note that the Bare Self-Attention works the best with tokens that are $2\times2$ patches of pixels, while the Transformer Block works better with $4\times4$ patches.

\begin{figure}
    \centering
    \includegraphics[width=1\linewidth]{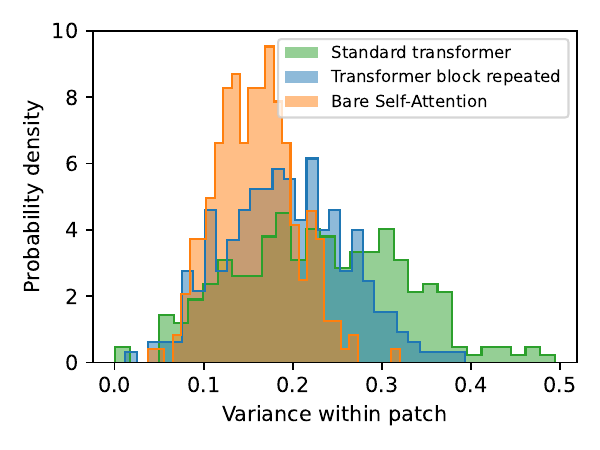}
    \caption{\textbf{Bare Self-Attention predicts more uniform patches.} We plot the distribution of variances of predicted pixels for different models, on the masked prediction task. As we make the model simpler, the variance within a patch decreases, meaning that the prediction becomes more and more uniform. For this plot, the Bare Self-Attention was trained on $4\times4$ patches like the other two models.}
    \label{fig:variances}
\end{figure}

To better understand the difference in the performance of the Transformer Block and the Bare Self-Attention, we measure the variance of predicted pixels within patches in the masked prediction task: from fig.~\ref{fig:variances} we see that the variance within patches becomes lower when we go from a full transformer to a transformer block to the Bare Self-Attention. In fact, we can see from the examples in fig.~\ref{fig:results} that the Bare Self-Attention tends to assign the same color to all the pixels in a patch and struggles with borders.

Therefore, it appears that the MLP helps in modulating the signal within the patches, which might explain why the Bare Self-Attention needs smaller patches.

\section{Conclusions and discussion}
\label{sec:discussion}
In this work we wrote the update of Self-Attention layers as derivatives of local energy contribution. To do this, we made the assumption that the attention weights can be written with an explicit dependence on the token positions, to model the effect of the positional encoding.
The local energy that we identified appears to be similar to a (negative,log) pseudo-likelihood, which inspired us to train the model by just minimizing the energy on training examples.
The trained model shows predicting capabilities consistently across two tasks. In particular, the model "stores" test examples as transient states of the dynamics. 
We also showed that a single transformer block used as a recurrent network share the same property, but has better reconstruction properties.

Even if the Bare Self-Attention recurrent network has worse reconstruction capabilities than whole transformers blocks, its simplicity has some conceptual advantages. First, the way in which we wrote the dynamics in eq.~\ref{eq:bareSA_dynamics} is much closer to physics than similar works on this topic. In particular, eq.~\ref{eq:bareSA_dynamics} resembles the update computation of a local field modulated by an attention mask, suggesting that further studies on attractor networks of vector spins might be desirable.
Another interesting point is that the training that we designed appears to be \emph{task-agnostic}, meaning that it produces a dynamics with the similar attractors and transient states regardless of the way in which we inject noise on the data.
For these reasons, and because we avoided trainable embeddings and MLPs, this approach may be theoretically tractable in the context of attractor networks. The results in this works are encouraging, given that the same concept of memories as transient states also appears in Vector Hopfield Networks with symmetric couplings \cite{nicoletti2024vector}.

On the numerical side, it would be interesting to find a way to include trainable embeddings and/or MLPs in the pseudo-likelihood training, to see if this approach can be extended to more complicated datasets, possibly providing a new scheme for training recurrent networks. 

\paragraph*{Code availability} The notebooks for the codes used in this work are available at \url{https://github.com/Francill99/self_attention_attractor_network}.

\section{Appendix}
\subsection{Architecture of the Self-attention recurrent net}
\label{sec:SA_architecture}
The input image is splitted into non overlapping square patches (indexed by $i$), as originally proposed in the vision transformer \cite{dosovitskiy_image_2021}. Given the number $a$ of (single channel) pixels in a patch, we map those pixels via an invertible transformation $A$ to a $2a$ dimensional normalized vector spin, as follows. Given the value of a pixel $p_k\in[0,1]$, we define the vector for pixel $k$ as $\mathbf{p}_k=(p_k,1-p_k)/\sqrt{p_k^2+(1-p_k)^2}$. Then we define the vector $\mathbf{s}_i$ representing the patch $i$ as the concatenation of $\mathbf{p}_k$ for $k\in \mathrm{Patch}[i]$. The transformation is the same for each patch, so no positional information is encoded in the spin, at variance with the vision transformer. 
Finally, the embdedded tokens $\mathbf{x}_i$ are computed with a linear transformation $F\in \mathbb{R}^{d\times 2a}$ applied to each vector spin: $\mathbf{x}_i = F \mathbf{s}_i$, 
where we choose $d\geq 2a$ to overparametrize the model. 
To compute the output images, the de-embedding operation is simply obtained using $F^{-1}$ and $A^{-1}$.
Since we do not want trainable embeddings, we select a fixed random orthogonal transformation: the columns of $F$ are chosen from a random orthonormal basis of $\mathbb{R}^d$, and then are multiplied by $1/\sqrt{a}$ so that $\mathbf{x}_i \mathbb{J}_{ij} \mathbf{x}_j$ does not depend on $d$ or $a$.
Finally after each iteration $\mathbf{x}_i^{(t+1)} = \sum_{j(\neq i)} \alpha_{i\gets j} \mathbb{J}_{ij} \, \mathbf{x}_j^{(t)} + \gamma \mathbf{x}_i^{(t)}$ each spin is normalized through layer-normalization, as described in \cite{vaswani_attention_2017}.

\subsection{Training of the Self-attention recurrent net}
\label{sec:SA_training}
The cost function in Eq. \ref{eq:e_i} 
can be trivially minimized by maximizing the of norm of weight matrices $\mathbb{J}_{ij}$. To avoid this, 
weights are initialized in $[\frac{-1}{2d},\frac{1}{2d}]$ and during training the L2 norm of the matrix is kept fixed. We control the magnitude of the interactions by manually multiplying $\mathbb{J}_{ij}$ with a scalar $\lambda$. We obtain the best results by setting $\lambda=5$ during the training and $\lambda=1$ during the evaluation on the test set.
We also keep $\mathbb{J}_{ii}=\bf{0}$.
A fundamental operation during training of transformers is the clipping of the gradient: in the SGD algorithm, when computing the error derivative of the $e_i$ cost in Eq. \ref{eq:e_i}, the presence of exponentials creates numerical instabilities, that are regularized imposing a maximum threshold of the L2 norm of gradients.

We train the Self-attention recurrent net for 20 epochs with a mini batch of 32.

\subsection{The Transformer block repeated}
\label{sec:block_training}
For this model we use single transformer block from the original vision transformer. Here we sketch the architecture, and for a detailed description we point the reader to the original reference \cite{dosovitskiy_image_2021}. 
The training is performed with backpropagation as usual, but instead of $\tau$ different layers the same transformer block is repeated $\tau$ times. Namely, the embedding transformation $F$ is a trained linear transformation (the de-embedding transformation is a different, trained $F_\mathrm{out}$ linear transformation); positional encoding is added to vectors after embedding; and each repetitions consists of a layer-normalization (LN) \cite{ba2016layernormalization}, followed by a self-attention (SA) operation as in Eq. \ref{eq:actual_SA}, a second LN and then the application of a multyilayer perceptron (MLP): the single transformer block can be summarized as
\begin{align}
    & u_i^{t+1} = \text{SA}(\text{LN}(x^t))+x_i^t
    \\& x_i^{t+1} = \text{MLP}(\text{LN}(u_i^{t+1}))+u_i^{t+1}
\end{align}
This block is iterated $\tau$ times. Critically, the value of $\tau$ used during training introduces a timescale in the behaviour of the model. We used the same choice as in \cite{jumper_highly_2021}: instead of training with a single $\tau$ value, at each training step we sample $\tau \sim \mathrm{Unif}([3,7])$. After last repetition we subtract the positional encoding from the embeddings before computing the output image.

We trained the transformer block for 100 epochs with a mini batch of 256. We used standard choices for the rest of the hyperparameters and the precise values can be found in the code available at  \url{https://github.com/Francill99/self_attention_attractor_network}. We used the same choices also for training the whole vision transformer.

\bibliographystyle{unsrt}
\bibliography{references}

\end{document}